\documentclass{article} 
\usepackage{iclr2023_conference,times}
\usepackage{graphicx}

\usepackage{amsmath,amsfonts,bm}

\def\eqref#1{equation~\ref{#1}}

\def\1{\bm{1}}

\DeclareMathAlphabet{\mathsfit}{\encodingdefault}{\sfdefault}{m}{sl}
\SetMathAlphabet{\mathsfit}{bold}{\encodingdefault}{\sfdefault}{bx}{n}

\usepackage{hyperref}
\usepackage{url}

\title{Adaptive Interventions for Global Health: \protect\\A Case Study of Malaria}

\author{África Periánez  \\
benshi.ai\\
Barcelona, Spain\\
\texttt{africa@benshi.ai} \\
\And
Andrew Trister \\
Bill and Melinda Gates Foundation \\
Seattle, USA\\
\texttt{andrew.trister@gatesfoundation.org} \\
\And
Madhav Nekkar \& Ana Fernández del Río  \\
benshi.ai\\
Barcelona, Spain\\
\texttt{\{madhav,ana\}@benshi.ai} \\
\And
Pedro L. Alonso \\
Faculty of Medicine, Universidad de Barcelona \\
Barcelona. Spain\\
\texttt{alonso@ub.edu}
}

\iclrfinalcopy 
\begin{document}

\maketitle

\begin{abstract}

Malaria can be prevented, diagnosed, and treated; however, every year, there are more than 200 million cases and 200.000 preventable deaths. Malaria remains a pressing public health concern in low- and middle-income countries, especially in sub-Saharan Africa. We describe how by means of mobile health applications, machine-learning-based adaptive interventions can strengthen malaria surveillance and treatment adherence, increase testing, measure provider skills and quality of care, improve public health by supporting front-line workers and patients (e.g., by capacity building and encouraging behavioral changes, like using bed nets), reduce test stockouts in pharmacies and clinics and informing public health for policy intervention. 

\end{abstract}


\section{Introduction}\label{sec1}
Not everyone benefits equally from the same treatment, has the same environment, or can receive therapy at the same time. Adaptive interventions consider human differences as treatment begins and during its course to ensure each person gets the treatment that continuously works for them. Interventions are adaptive in that they adapt to the individual (providing personalization) and their evolving context and needs (i.e., adjusting dynamically).

Both clinical and patient actions are essential to high-quality, low-cost, effective healthcare. Human behavior constitutes the primary mode for activating health improvements, including decisions made by clinicians and their patients. In medical practice, clinicians can be gently nudged to improve decision-making by providing a decision architecture in which optimal default clinical actions are
suggested. Outside controlled clinical environments (e.g., intensive care units), patient actions (e.g., lifestyle choices and treatment adherence) primarily determine health outcomes.
Mobile health is key to promoting continuous and proactive healthcare monitoring. It can, for example, mitigate the lack of medication adherence, which can be low and challenging to address, as patient actions control healthcare delivery outside medical environments.


The data generated by the users of digital applications are instrumental in determining past and current behaviors and predicting future conduct. This information can be used to deliver personalized mobile-mediated interventions. Such interventions aim to provide the appropriate type or degree of support by adapting to an individual’s changing internal and contextual states~\citep{Nahum2018, Menictas2019, Carpenter2020, Bidargaddi2020, Coppersmith2021}.

Mobile health applications also serve as a direct channel of communication with their users, from front-line healthcare workers to patients and the general public. Interventions and incentives can be delivered directly to the users through their phones. Machine learning (ML) can help generate predictions regarding the behavior of app users, health outcomes, and their contexts. These predictions can be integrated with real-time information pertaining to the users’ choices and circumstances to determine the individuals or groups needing additional support. Moreover, these predictions can be used to personalize the timing of the interventions delivered to each user. Similarly, reliable predictions regarding the evolution or variations in the demand for different medical supplies can help establish reminder and suggestion systems that can help ensure that all the supply chain actors (e.g., pharmacists) maintain adequate stocks of essential supplies at all times. Motivational prompts, personalized reminders, incentives, engaging elements (such as game-designed elements), and other nudges can be used to boost and reinforce good practices~\citep{Hrnjic2019}.

Adaptive interventions in healthcare have primarily been deployed in resource-rich countries~\citep{Nahum2018, Hardeman2019}. Given the immense disease burdens borne by low- and middle-income countries and the increasing smartphone penetration in these regions, the use of adaptive interventions to improve health outcomes may be highly beneficial to global health. In this research, we explore this potential based on a case study of malaria. A discussion of the ML methods can be found in the appendix \ref{appendix:maths}.


\section{Malaria: A Case Study}
\label{sec:malaria}
Malaria, caused by the parasite \emph{Plasmodium falciparum}, remains a deadly illness across much of the Global South.  

Malaria in pregnancy and infancy is 
a major public health concern and a key driver of maternal and newborn mortality~\citep{Tarning2016}. In 2020, nearly 630,000 deaths worldwide were caused by malaria, a disproportionate share of which (96\%) occurred in Africa~\citep{WHO_malaria_factsheet_2022}. In Sub-Saharan Africa, malaria is responsible for 12\% of all child fatalities~\citep{RoserRitchie2019}, and children under five account for approximately 80\% of all malaria deaths ~\citep{WHO_malaria_factsheet_2022}.\looseness=-1

Although global malaria incidence and mortality have substantially decreased in the recent two decades, progress stalled since 2015~\citep{Noor2022}, and researchers worry that the COVID-19 pandemic has disrupted malaria intervention coverage, reversing these gains ~\citep{Weiss2021}. 

\subsection{Surveillance: Progress Towards Elimination} 
Even with the World Health Organization's (WHO) historic approval of a malaria vaccine~\citep{Alonso2022b}, senior leaders caution that a broader approach will be needed for malaria eradication, including improved collection and usage of high-quality data---from health-management information systems and electronic databases to geospatial models~\citep{alonso2021}---and flexible evaluation and implementation of interventions by local decision-makers~\citep{Alonso2022}. According to a global landscape review, malaria surveillance systems in 2015–-2016 were ``insufficient to support the planning and implementing of targeted interventions and measure progress toward malaria elimination''~\citep{lourenco2019}. Such elimination efforts require the accurate notification of individual cases within 24 h of diagnosis to provide timely and targeted responses, which essentially represent adaptive interventions~\citep{who2015}. 

Notably, such malaria surveillance frameworks must be integrated, data-driven, tailored, and based on mobile platforms. Mobile health for malaria surveillance, that uses a combination of message- and application-based reporting, can support health workers and clinicians in recording malaria case information~\citep{githinji2014,baliga2019,RTI2020,oo2021,bhowmick2021}, decreasing delays in case reporting to health officials, and improving the quality of data collection. Data fragmentation remains a barrier to a cohesive malaria response, and previously siloed data from diverse sources (private/public healthcare providers, government agencies, etc.) must be integrated to
create 
a real-time, case-based malaria surveillance system~\citep{rahi2020}. Automation is essential, as analytics related to threat monitoring, requirement identification, and system performance must be readily available for decision-makers~\citep{ohrt2015}.  

A comprehensive malaria surveillance system can inform policy interventions, e.g., the allocation of mosquito nets, tests, and antimalarials can be targeted to favor individuals and communities in need. 
By understanding the changes in people’s mobility and clustering, the impact of non-pharmaceutical interventions can be evaluated, and geographic areas in which additional actions may be helpful can be identified~\citep{grantz2020}. Several studies revealed that the characterization of travel patterns through geolocation data and their combination with contextual information on malaria incidence can
inform strategies to target travelers and 
reduce transmission~\citep{milusheva2020}.

\subsection{Prevention}
For pregnant women visiting institutions for antenatal care, the WHO recommends the administration of intermittent preventive treatment with sulfadoxine--pyrimethamine and distribution of long-lasting insecticide-treated mosquito nets (LLINs)~\citep{salomao2017}. 

Unfortunately, many pregnant women in low- and middle-income countries do not receive either intervention owing to stock-outs in data-fragmented health systems~\citep{salomao2017}. Accurate demand forecasts that can integrate real-time contextual information are needed to ensure equitable and efficient allocation of important preventive goods. For example, by building geospatial models and combining data from net manufacturers, national programs, and cross-sectional household surveys, researchers can develop detailed maps of LLIN access and usage~\citep{Bertozzi2021}. These analyses can also be automated to identify geographic areas in which additional supplies or different actions may be required.   

Patient behavior also determines the efficacy of the interventions, as many pregnant women fail to take their pills~\citep{mubyazi2005} or do not use bed nets for various practical reasons~\citep{manu2017,gultie2020}. These aspects highlight the importance of communication to encourage behavioral changes~\citep{ricotta2014}.

For children aged 3–-59 months, WHO recommends seasonal malaria chemoprevention (SMC) during the months of peak malaria transmission. Although SMC effectively controls malaria and reduces hospitalizations~\citep{diawara2017,baba2020,issiaka2020,cairns2021}, clinical data and pharmacokinetic analyses reveal that complete adherence to treatment is observed in fewer than 20\% of children outside the study setting~\citep{ding2020}. Targeted interventions  for behavioral changes are required to improve the real-world effectiveness of SMC.

Caregiver-targeted message-based interventions to increase preventive health behaviors (such as sleeping under a net) have been noted to be successful in decreasing the malaria prevalence in children under the age of five~\citep{mohammed2019}. Despite the vast potential of mobile health solutions, one-size-fits-all interventions are typically implemented in which a standard message is sent to all participants. There exists an enormous opportunity for the delivery of personalized interventions that can appropriately incentivize a given user and guide targeted public health campaigns, which may include gamification and leveraging of individual social networks~\citep{ernst2017}. As a potential example, researchers recently tested an ML model in conjunction with an accelerometer-based approach for measuring a range of LLIN use behaviors: Although these technologies represent a proof of concept at present, they can support the implementation of financial incentives based on granular LLIN-use monitoring over longer time periods~\citep{Koudou2022}. 

The only way to understand which interventions work best and which incentives drive behavioral change is by running experiments on the ground~\citep{banerjee2010,bates2012, Zhou2020b, Zhou2021}.

\subsection{Quality of Care} 
\subsubsection{Diagnosis}  
Malaria is one of the most under-diag\-nosed and over-treated diseases. The potential for severe outcomes means that any patient with a fever (especially a child) may be administered treatment for malaria, often without having received a diagnostic test~\citep{Amankwa2019, Ajibaye2019, boadu2016, beisel2016}. The administration of drugs without a conclusive test may accelerate antimalarial resistance, an increasingly worrying problem according to WHO malaria experts~\citep{Rasmussen2022}. 

Although microscopy is the gold-standard for malaria diagnosis, its implementation remains infeasible in many resource-constrained settings~\citep{beisel2016}. Consequently, many stakeholders have turned to rapid diagnostic tests (RDTs). However, their access remains limited, with frequent stock-outs~\citep{boadu2016,blanas2013} that may be related to inaccurate record-keeping. Digital tools and demand forecasting algorithms can alleviate this problem, as has been proved in India with an app that included a supply chain management component~\citep{rajvanshi2021}.

Furthermore, many lay community health workers struggle to appropriately perform RDTs~\citep{boadu2016,blanas2013, beisel2016}, and thus, capacity-building efforts to increase their skills are urgently required. Digital monitoring of the health workers' performance can serve as an effective quality control strategy and a mode of delivering feedback~\citep{Laktabai2018}. Personalized digital nudges can direct the health workers who need additional support to specific online learning resources or encourage them to sign up for an in-person training session. 

Patient-facing steering may also be required, as sick patients who receive a negative test may still expect treatment. Patient and provider education around antimicrobial resistance 
may help, in addition to increasing awareness and clinical decision support for the management of other febrile illnesses. 


\subsubsection{Artemisinin Combination Therapy (ACT)}
ACT is the first-line malaria treatment throughout most of the malaria-endemic world. Notable issues include stock-outs~\citep{blanas2013, rowe2009,oconnell2011} and the distribution of low-quality antimalarials, especially in urban areas~\cite{newton2017}. Most antimalarials are distributed through the private sector, where non-artemisinin therapies are prevalent and 5--24 times less expensive than quality-assured ACT~\citep{oconnell2011}. In this context, it is important to strengthen the supply chain with a data-driven approach, potentially by using demand forecasting algorithms to send appropriately timed reminders to pharmacists to ensure that they restock necessary supplies.\sloppy

The quality of care and clinical management of malaria remains widely varying and substandard. In Sub-Saharan Africa, less than one-third of the children diagnosed with malaria receive both a blood test diagnosis and appropriate antimalarial treatment~\citep{cohen2020}. Evidence for gaps in the community health workers' and drug dispensers' ability to appropriately manage malaria has been found in several countries~\citep{rowe2009,blanas2013,chowdhury2020, buabeng2010,kamuhabwa2013}. Capacity-building efforts are clearly necessary, and adaptive interventions can facilitate the assignment of appropriate content to each  worker~\citep{Katsaris2021}. AI-based user segmentation and behavioral phenotyping can yield highly granular cohorts to be focused on. For example, we research the capacity development of midwives by predicting their demand for specific contents~\citep{guitart2021midwifery} 
and building recommendation systems for an e-learning app by predicting the chance that a given user clicks on a certain item~\citep{Guitart2021}. 

Lastly, patient adherence to antimalarial regimes is a notable issue, with estimates suggesting that it ranges from 40\% to 65\%~\citep{amponsah2015,mace2011,onyango2012}. Trials of message-based-reminders to healthcare workers~\citep{zurovac2011,kaunda2018} and patients~\citep{macias2022} have shown promise while highlighting the necessity of personalized interventions involving relevant and actionable messages~\citep{buabeng2010}. In the case of tuberculosis, where adherence is also a problem, researchers have noted that an ML-driven approach to predict TB adherence risk in conjunction with adaptive interventions dynamically increases the odds ratio of next-day treatment adherence verification by 35\%~\citep{Boutilier2021}.  

\subsubsection {Pre-Referral Treatment}
Another crucial area where malaria care delivery can be strengthened is pre-referral treatment, usually rectal artesunate (RAS), recommended by WHO for young children with suspected severe malaria when injections are not available~\citep{who2022}. Although the effectiveness of RAS has been demonstrated in placebo-controlled trials~\citep{gomes2009}, more recent studies have indicated that its introduction in real-world conditions may increase the malaria case-fatality rates~\citep{who2022}. This phenomenon is attributable to several reasons, such as low patient adherence to referral guidelines~\citep{simba2010} and inadequate ability of health workers to deliver pre-referral care~\citep{amboko2022} appropriately. Patient- and provider-facing steering and capacity building are needed to strengthen health systems and ensure that guidelines are followed.

\section{Conclusions}
\label{sec:conclusions}
Combating the devastating toll of malaria requires multifaceted and innovative strategies. Technology alone is not a panacea, and no simple solution exists. However, ensuring high-quality data collection from diverse sources, such as mobile phones, supply chains, public surveys, and electronic health records, can play a crucial step. These data can be integrated to obtain intrinsic and contextual information that can drive personalized and adaptive interventions. 


The use cases are vast: Reminding antenatal care clinics to stock up on LLINs; providing capacity-building resources to struggling community health workers; supporting a district health officer in responding to a new malaria outbreak; incentivizing pharmacists to administer diagnostic tests before prescribing treatments; or encouraging patients to adhere to a treatment regime. 







\subsection*{Acknowledgments}

This work was supported, in whole or in part, by the Bill \& Melinda Gates Foundation INV-022480. Under the grant conditions of the Foundation, a Creative Commons Attribution 4.0 Generic License has been assigned to the Author Accepted Manuscript version that might arise from this submission.

\bibliography{iclr2023_malaria}
\bibliographystyle{iclr2023_conference}

\appendix
\section{Appendix: Tools and Methods}
\label{appendix:maths}

This appendix describes the proposed conceptual and methodological framework: a data-centric behavioral ML platform~\citep{Tang2021} that leverages logs from different types of mobile health solutions---together with contextual information---to deliver adaptive interventions to healthcare workers and their patients directly through their phones.


\subsection{Data-Centric Platform}


Our platform is data-centric in that data tracking, and labeling lies at its core. Integration with the different apps is achieved through our \emph{Software Developer Kit} (HealthKit SDK), which 
provides specifications on what information should be logged (and how) and the messaging service that delivers the interventions. 
Integration through this SDK ensures that
various user metrics and traits characterizing engagement and behavior are readily available, 
which can help clarify the individuals to be targeted by specific interventions and the time for delivering the interventions.

\subsection{Machine Learning}
The application of data science and ML methodologies to extract and predict valuable information to inform intervention design lies at the core of the software we build. As is generally the case, no single model is best across all datasets and use cases. Use-case-specific data pipelines transform incoming information through the HealthKit SDK into metrics ready to be consumed by the statistical and ML models, which in turn produce additional metrics that can be used in the intervention definition.

\subsubsection{Reinforcement Learning}
Reinforcement learning (RL) is the ideal paradigm for sequential decision-making in dynamically evolving contexts that respond to those decisions. It allows us to continually improve how we make choices for a patient at any given moment, maximizing the potential for positive outcomes while minimizing undesired side effects. A competition on policy learning for malaria control using RL was for example included as part of the KDD Cup Challenge 2019~\citep{Zhou2020a, Nguyen2019, Zou2021}.

Contextual are a formulation of the RL problem with limited state representation, and dynamics \citep{Burtini2015, Yao2021, Dwivedi2022, Zhang2022}. Similarly, restless bandits can be used for resource allocation~\citep{Mate2022}. They are significantly less data-intensive than other RL approaches while being robust and tractable. This framework is instrumental in the context of adaptive intervention delivery. 
At their core lies the exploration-exploitation trade-off, i.e., the compromise between clinical research (to gather knowledge about treatments) and clinical practice (to benefit the patients) by assigning the best intervention possible based on all available information at that point.


\subsubsection{Time-varying and Dynamic Prediction Modeling}
Time series datasets such as clinical records represent valuable sources of information sometimes spanning a patient’s entire lifetime of care.
The approach to decision-making described in this paper is dynamic (i.e., sequential and adaptive). The analytic and predictive modeling to support this needs to be similarly dynamic, with the time-dependent evolution of the systems of interest and their characteristics at its core. Supervised and unsupervised ML with time-varying data (such as the survival analysis described in the next section), time series modeling, and longitudinal data processing should be part of the toolbox. We can use them to understand and predict how the systems of interest behave and evolve to inform our decision-making.

For behavioral nudging, besides the individual predictions generated using survival analysis described below, multivariate time series forecasting can be critical to optimizing specific interventions, such as reminders to prevent stockouts of medical supplies based on demand prediction. Furthermore, considering the time-varying nature of data and interventions within an RL framework allows us to understand and optimize a patient’s treatment as different sequential interventions in time instead of as a single-point decision.

Predictive modeling 
can help define the target individuals for the intervention and appropriate delivery time. 
As discussed, accurate demand prediction is key to optimizing the supply chain and inventory. Forecasting methods that can learn simultaneously from multiple time series, often combining deep and state space modeling elements, have been noted to be effective~\citep{Seeger2017, Salinas2019, Lim2019, Salinas2020, Benidis2020}.

\subsubsection{Deep and Ensemble Survival Analysis}

With time dependence at its core, survival analysis refers to a collection of algorithms used to predict time to an event of interest (which was traditionally death or organ failure)~\citep{Wright2017, Fu2016, Lee2020}. These methodologies can be used to predict behavior and health outcomes at the individual level, as they are well suited to extract information from censored data (i.e., models that can learn from subjects who have yet to experience the event of interest besides those that have). Their output is a survival curve for each subject, indicating the probability of not having experienced the event of interest depending on the time. By carefully selecting the events we predict and how we measure time, we can profile individuals using multiple predictions and risk scores (of clinical complications or of lack of adherence to treatment, for example).


\subsubsection{Recommendation Systems}

Recommendation algorithms 
can be used
to decide the techniques to target health workers, patients, and the general public. 
Different 
methodologies are expected to be suitable for different apps and use cases, 
from deep learning-based click-through-rate predictions~\citep{Qu2016, Guo2017, Lian2018, Lu2020, Guitart2021} to collaborative, interactive recommendation systems~\cite{Ie2019, Liu2019, Lillicrap2019, Chen2020}. 
We foresee a growing shift toward reinforcement-learning-based methods beyond the (relatively widespread) use of contextual bandits~\citep{Barrazaurbina2020}. This approach is particularly useful
for optimizing a \emph{sequence of interventions} while reconciling short- and long-term goals in the desired outcome. 

\end{document}